\documentclass[conference,letterpaper]{IEEEtran}  
\usepackage[letterpaper, total={7in, 9in}]{geometry}
\usepackage{cite}
\usepackage{amsmath,amssymb,amsfonts}
\usepackage{algorithmic}
\usepackage{graphicx}
\usepackage{caption}
\usepackage{textcomp}
\usepackage{xcolor}
\usepackage{tikz}
\usetikzlibrary{fit,positioning}
\usepackage{listofitems} 
\usepackage{pgfplots}
\usepackage{multirow}
\usepackage{pgfplots, pgfplotstable}
\usepackage{filecontents}
\usepackage{epsdice}
\usepgfplotslibrary{polar}
\tikzstyle{mynode}=[thick,draw=blue,fill=blueue!2020,circle,minimum size=15]
\def\BibTeX{{\rm B\kern-.05em{\sc i\kern-.025em b}\kern-.08em
    T\kern-.1667em\lower.7ex\hbox{E}\kern-.125emX}}
\DeclareMathOperator*{\argmax}{argmax}

\pgfplotsset{compat=1.16}

\begin{document}

\pagenumbering{gobble}

	\addtolength{\oddsidemargin}{0.75mm}
	\addtolength{\evensidemargin}{1.5mm}
	\addtolength{\textwidth}{-1.5mm}

	\addtolength{\topmargin}{2mm}
	\addtolength{\textheight}{8mm}

\title{Learning Object Manipulation With Under-Actuated Impulse Generator Arrays}

\author{\IEEEauthorblockN{C. Kong}
\IEEEauthorblockA{Mitsubishi Electric Research Labs\\
Cambridge, Massachusetts, 02139\\
email: kczttm@gmail.com}
\and
\IEEEauthorblockN{W. S. Yerazunis}
\IEEEauthorblockA{Mitsubishi Electric Research Labs\\
Cambridge, Massachusetts, 02139\\
email: yerazunis@merl.com \\
\it{correspondence author}}
\and
\IEEEauthorblockN{D. Nikovski}
\IEEEauthorblockA{Mitsubishi Electric Research Labs\\
Cambridge, Massachusetts, 02139\\
email: nikovski@merl.com}}

\maketitle
\thispagestyle{plain}
\pagestyle{plain}

\begin{abstract}
For more than half a century, vibratory bowl feeders have been the standard in automated assembly for singulation, orientation, and manipulation of small parts. Unfortunately, these feeders are expensive, noisy, and highly specialized on a single part design bases. We consider an alternative device and learning control method for singulation, orientation, and manipulation by means of seven fixed-position variable-energy solenoid impulse actuators located beneath a semi-rigid part supporting surface. Using computer vision to provide part pose information, we tested various machine learning (ML) algorithms to generate a control policy that selects the optimal actuator and actuation energy. Our manipulation test object is a 6-sided craps-style die. Using the most suitable ML algorithm, we were able to flip the die to any desired face 30.4\% of the time with a single impulse, and 51.3\% with two chosen impulses, versus a random policy succeeding 5.1\% of the time (that is, a randomly chosen impulse delivered by a randomly chosen solenoid). 
\end{abstract}

\begin{IEEEkeywords}
Optimal control under uncertainty, stochastic modeling, learning control
\end{IEEEkeywords}

\section{Introduction}
Automated assembly of products makes use of various factory automation devices whose purpose is to put the component parts together in the correct order and position. When typical first-generation industrial robots are used for the actual assembly, they execute the exact same sequence of operations without any variation. The only way this would be successful is if the component parts are presented in the exact same position and orientation, and it is the job of other types of factory automation equipment to make sure that this is the case. A very common and popular such device is the vibratory bowl feeder (VBF) \cite{Sgriccia1950Feeder2654465}  that uses a circular vibratory pattern and a specially designed ramp to bring parts up the ramp in the desired orientation.

VBFs are typically noisy, expensive, and difficult to design, due to their size and complexity. With costs reaching hundreds of thousands of dollars and lead times of three to six months, they make economic sense only for very large production runs, and are a poor match to the increasing trend towards high-mix, low-volume manufacturing. A new generation of industrial robots equipped with cameras has made it possible to grasp parts in a range of orientations, as long as they are sufficiently singulated from one another. This has led to the emergence of simplified part feeders where the parts are deposited not in a bowl, but on a flat surface which vibrates in a fixed pattern, eventually singulating at least some of the parts so that they can be grasped by a camera-equipped robot. This solution reduces drastically the noise, size, and cost of the feeder, as the vibration pattern is generic and no custom design is needed for each part. 
Still, this solution does not eliminate the problem of having the part often lie on the wrong facet. The robot has some flexibility about how to grasp the part, but at best the robot can approach it from a direction in no more than half of the unit sphere, that is, from above. To deal with this, when the part is facing the wrong way up, the robot would have to pick it up, place it down on a different facet, and regrasp it. There is no generic robot program to do that reliably for an arbitrary part geometry, so a customized program would need to be developed. Moreover, even if such a program were developed, the robot would have to spend time executing it, instead of doing actual assembly, thus increasing the takt time of the assembly operation, which is highly undesirable. 

\begin{figure}[h!]
    \centering
    \includegraphics[width=0.4\textwidth]{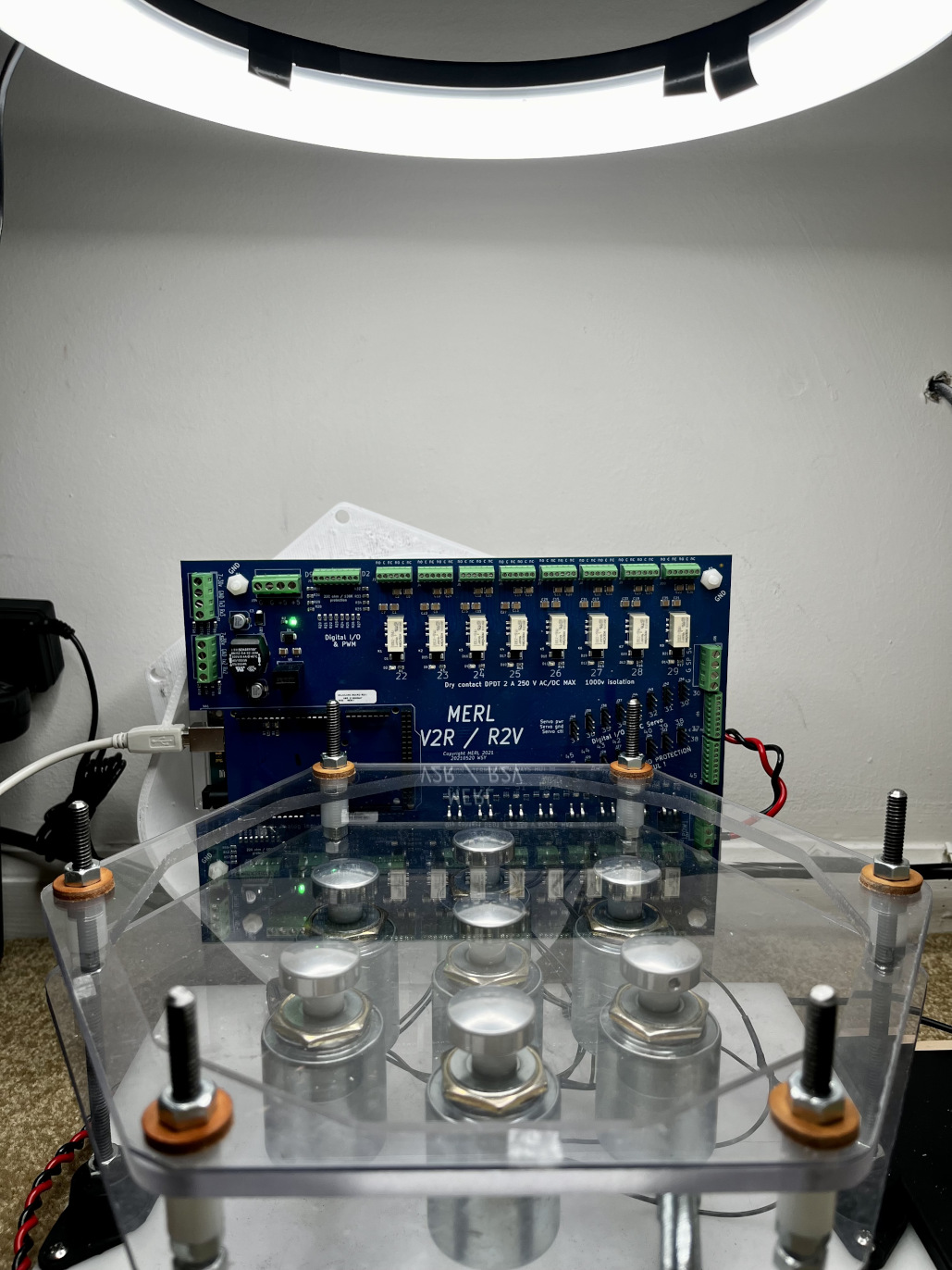}
    \caption{\ The 7-solenoid impact manipulation surface "Thumper"; the seven solenoids beneath the transparent support surface are in the foreground, with the controller board in the background}
    \label{clear_top_thumper}
\end{figure}

To solve this problem, we propose a novel design for a part feeder that uses a set of solenoids mounted under the surface that the parts have been placed on to impart impulse shocks onto the surface so as to flip the parts to a different facet, if the current one is not suitable for grasping. The device is equipped with a camera whose purpose is twofold: first, to recognize which facet the part is lying on, and second, to register the part's position and orientation in order to decide which solenoid to fire in order to maximize the chance of success in changing the facet. Note that this camera could be the same camera that the robot uses for grasping decisions, so it adds no additional cost to the system, while effectively making the part feeder adaptive.

This paper deals with the problem of deciding how to control the system in order to manipulate the parts in the feeder in an optimal way. The motion of the manipulated object involves complex contact dynamics that vary according to the geometry of the part, and traditional physical modeling would be prohibitively difficult and expensive. For this reason, we adopt the methodology of learning control by learning probabilistic models of the outcomes as a result of applied controls, and using them to choose the optimal control \cite{Atkeson1997LocallyControl}. Section \ref{sec:thumper} describes the design of the mechanism and its instrumentation with sensors, and Section \ref{sec:control} proposes a learning controller based on learned outcome models. Section \ref{sec:die} describes experimental verification with different control objectives, and Section \ref{sec:conclusion} concludes and proposes some directions for improving the success rate of the device and its controller.


	\addtolength{\topmargin}{-4mm}
	\addtolength{\textheight}{4mm}
	
\section{Design and Operation of the Experimental System}\label{sec:thumper}

Our experimental "smart bowl", called Thumper, is a seven-solenoid impulse-drive open-bowl manipulator. It is equipped with an HD webcam running at 30\,fps, and has individual control of the solenoid impulse generators, including the time and duration of the applied impulse.  The control software takes the video in, processes it with OpenCV to estimate the pose of the manipulated part, applies one of several ML methods to generate a policy for manipulating the object to a desired outcome, and from the generated policy and the observed current object pose, determines and issues impulse commands to impulsively maneuver the object into the desired state.

\begin{figure}[!h]
    \centering
    \begin{minipage}{0.22\textwidth}
        \centering
        \includegraphics[angle = 0, width=1\textwidth]{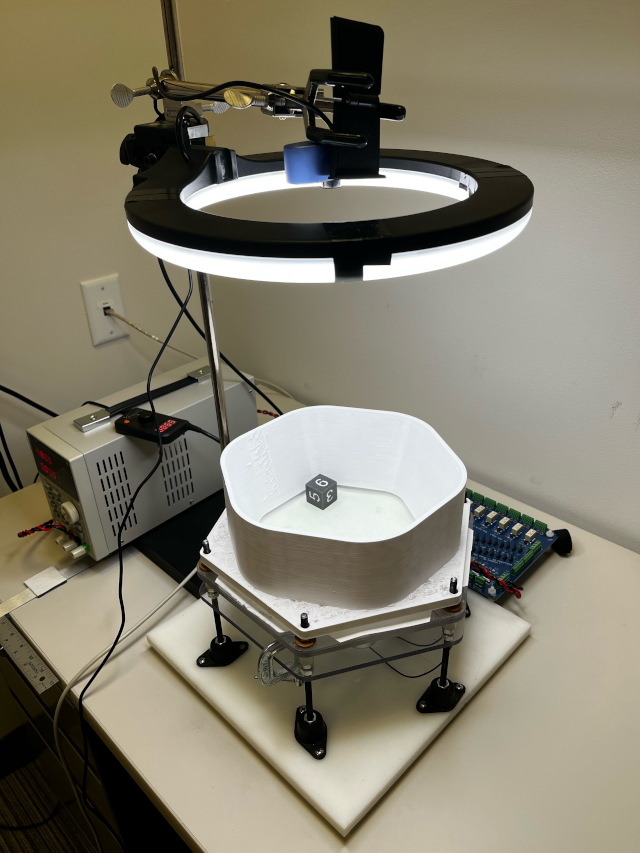} 
            \caption{Thumper in operation mode} \label{with_dice}
    \end{minipage}\hfill
    \begin{minipage}{0.22\textwidth}
        \centering
        \includegraphics[angle = 0, width=1\textwidth]{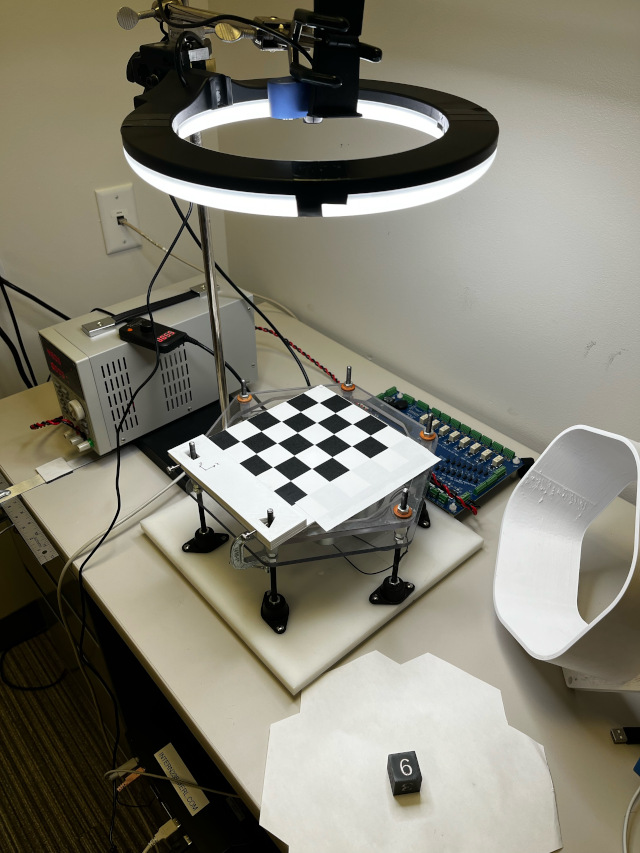} 
        \caption{Thumper in calibration mode}\label{cali}
    \end{minipage}
\end{figure}

The overall system can determine the pose of a $\sim$25 mm test cube with an accuracy and repeatability of slightly less than one millimeter in X and Y, and about one degree in rotation, as projected onto the horizontal plane.  The actual manipulation commands are fairly sparse; the system can only select which one of the seven solenoids to fire, and choose a firing duration.  The firing duration is limited by empirical observation to be between 8 and 25 milliseconds, as durations below eight milliseconds are observed to be inadequate to actually move the test object, and impulses over 25 milliseconds have no greater authority in moving a test object. To confine the parts onto the bowl floor, a white 3-D printed hexagonal "corral" is mounted ~5mm above the bowl floor. The mechatronic parts of the system can be seen in Figure \ref{clear_top_thumper} and the assembled system in Figure \ref{with_dice}.

A 1080p HD webcam mounted above the bowl  provides $\sim$30 frames per second to the computer vision (CV) system that locates the test objects and determines the test object pose.  To calibrate the camera, we use a  precise checkerboard mounted on a 3-D printed mount that positions directly against the vertical rods in a three-point kinematic arrangement, as shown in Fig. \ref{cali}.

\begin{figure}[h!]
    \centering
    \includegraphics[width=0.6\linewidth]{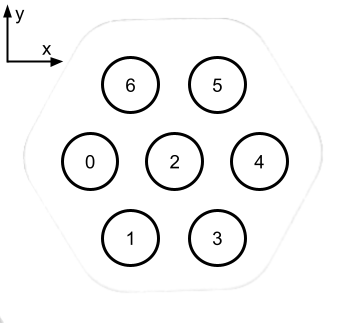}
    \caption{The geometrical layout of the solenoid array; the inter-solenoid spacing is 60mm.}
    \label{solenoid_layout}
\end{figure}

For naming convenience, in this paper we will use the term "thumper" to indicate the entire apparatus.  We will also use the term "thumper" with a number to indicate one of the seven sets of PowerFET, solenoid, and striker heads. 
The actual layout and numbering of the solenoids are shown in Figure \ref{solenoid_layout}.

\begin{figure}[!h]
    \centering
    \includegraphics[width=0.48\textwidth]{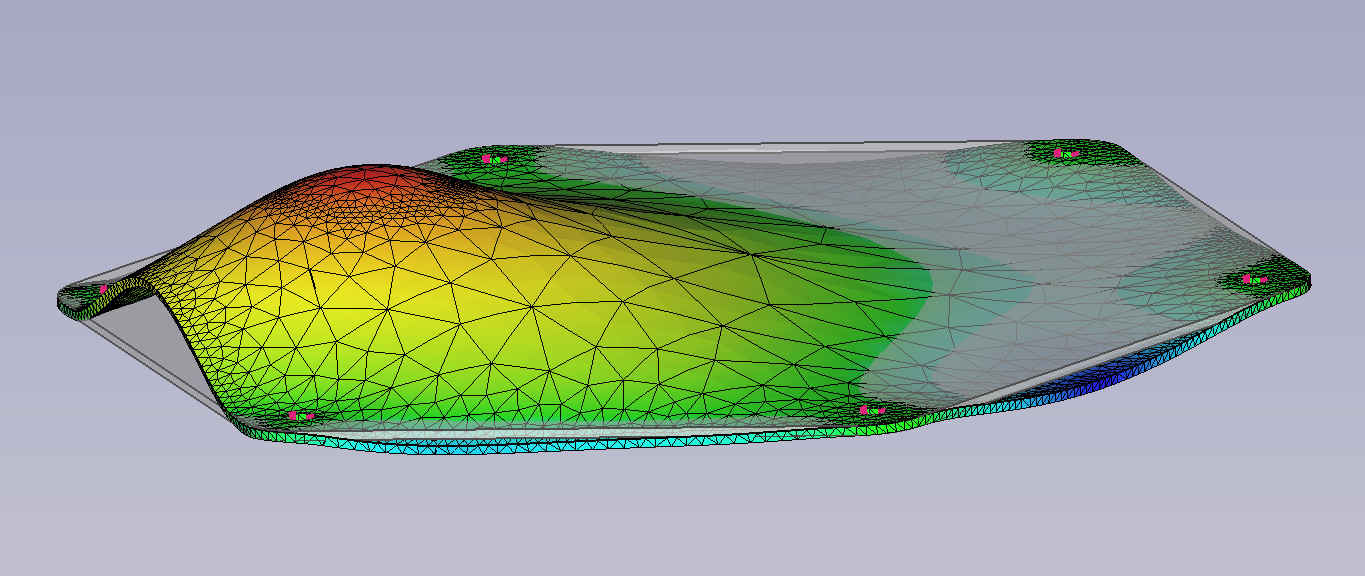}
    \caption{FEM of a single solenoid's static effect on the support surface; displacements accentuated 10x}
    \label{fem}
\end{figure}

\begin{figure}[!h]
    \centering
    \includegraphics[width=0.25\textwidth]{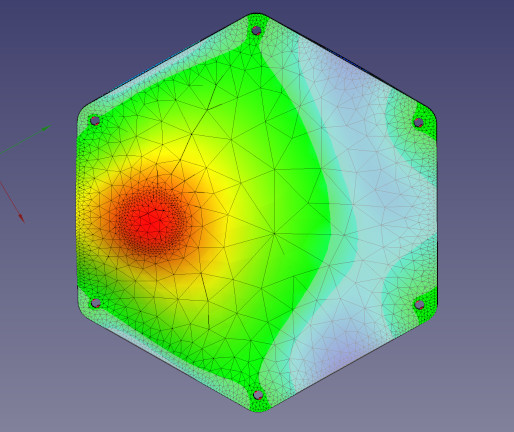}
    \caption{FEM top view showing Z displacement; the blue overlay shows areas where the support surface is moving downward.}
    \label{fem_topview}
\end{figure}

Figure \ref{fem} shows an FEM analysis of the bowl floor distorted by the static force of an impulse solenoid and Figure \ref{fem_topview} shows the top view on Z displacement alone; of note is that the areas of greatest Z-motion and the areas of greatest tilt are not the same, nor are they exact inverses of each other.  We surmise that this is actually a useful attribute, because to change the pose of the object, we must impart both a vertical impulse sufficient to get the object into the air, and also a rotational impulse sufficient to cause an adequate  rotation of the object before ground contact resumes.

\section{Learning Object Manipulation}\label{sec:control}
\begin{figure}
    \centering
    \includegraphics[width=0.3\textwidth]{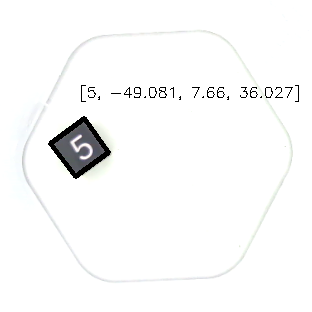}
    \caption{CV recognition of a state $(s, x, y, \theta)$ of the die: face=5, x=-49.801mm, y=7.66mm, angle=36.027 degrees}
    \label{die_with_states}
\end{figure}

We are interested in manipulating objects whose geometries allow them to stand stably on one of (relatively few of) their sides. Examples of such objects are hexagonal nuts and bolts, IC chips, etc. The state of such an object would be characterized by the side $s$ it is on (an integer), as well as its position $(x,y)$ and orientation $\theta$ (real numbers). The objective is to devise a control policy $u=\pi(s,x,y,\theta)$ that selects which solenoid $u_s$ to fire with duration $u_d$ so as to maximize the probability of moving the object into a desired state, $u=(u_s,u_d)$. This desired state can be described in terms of one or more of the state components, for example changing the face the part is lying on, or also possibly bringing it to a desired position and orientation. Let the Boolean function $g(s,x,y,\theta)$, provided by the user, indicate whether state $(s,x,y,\theta)$ is a desired goal state or not.

Modeling the effect of impulse shocks on the manipulated parts from the pertinent physical principles is usually extremely difficult. Modern physics engines implement the relevant physical laws of motion, as well as suitable contact models, and can generate and simulate the equations of motion automatically from geometrical scene descriptions, thus alleviating the need to create a dynamical model manually. However, this still involves a painstaking process of careful geometric calibration of the scene, possibly along with tuning a variety of contact model parameters, such as coefficients of friction, restitution, stiffness, surface roughness, edge and vertex radii, etc. A recent study on a task very similar to ours (rolling a cube on a flat surface) showed that even after very careful calibration, the behavior of the system was largely unpredictable and not very consistent across multiple physics engines \cite{Chung2016PredictableEngines}. This reflects the inherently chaotic dynamics of such systems, where rolling from facet to facet is associated with bifurcations in the system's dynamics. The bifurcation parameters include many of those of the contact model --- for example, whether a cube will roll to the next facet for a given angular and linear velocity or remain on the previous one would depend on how much its edge will slip on the surface, and that is determined by the friction coefficient; assuming an incorrect value for that coefficient could predict a very different outcome as to how the cube will land. Moreover, physics engines are inherently deterministic, as their purpose is to predict the one physical reality that will happen, whereas the chaotic dynamics of irregular rolling parts might be better modeled by stochastic models for control purposes.       

For these reasons, we adopt a learning control approach. There is a great variety of learning control methods in the literature, whose success largely depends on the nature of the control problem being solved. Early work on learning non-prehensile manipulation of parts by means of tilting a tray made use of observed examples of the effect of actions (in that case, the direction of the tray's tilt) to learn a model of these actions, and used this model for planning \cite{Christiansen1991LearningModels}. This method was based on earlier work on stochastic learning automata (SLA), \cite{Narendra1974LearningSurvey}, and discretized the state space of the manipulated part (position on the tray) into rather coarse regions. This matched well the usual assumption of SLA for relatively few discrete states, and made the learning problem tractable, but led to the introduction of additional uncertainty and stochasticity in the model due to partial state observability, on top of the already significant stochasticity of the system due to complex contact and impact dynamics. Although our control problem bears strong similarity to the one in \cite{Christiansen1991LearningModels}, we believe that an approach that does not quantize the state into a few coarse discrete states would be more productive. 

Another distinct approach to learning manipulation has been to learn a full state-space model of the system dynamics, using various system identification methods \cite{Ljung1997SystemUser}.  Whereas this approach has been very productive for linear systems, the complicated non-linear nature of contact dynamics has required the application of advanced methods for learning non-linear and possibly hybrid discrete/continuous dynamical models. Various universal function approximation methods have been used to learn system dynamics, and neural networks in particular have been investigated extensively for a long time \cite{Narendra1990IdentificationNetworks,Jordan1992ForwardTeacher,Narendra1992NeuralSystems}. Recent interest in model-based reinforcement learning has renewed research efforts to find good methods for learning world models \cite{Moerland2020Model-basedSurvey}. Recently proposed Contact Nets have improved considerably the accuracy of predictive models with respect to earlier dynamical models based on standard neural networks \cite{Pfrommer2020Contactnets:Representations,Parmar2021FundamentalDynamics}. However, learning such models is quite complicated, and might also be an overkill for our control problem, where prediction of the entire future trajectory of the manipulated part is not really necessary, and predicting the stable resting state would suffice. 

For this reason, we focused on learning predictive models that predict only the resting state of the manipulated part as a result of a particular action (solenoid fired). Similar to SLA, these predictive models are probabilistic, to capture the inherent stochasticity of the complex contact dynamics involved. However, unlike SLA, our models use the full continuous state of the manipulated part, measured as precisely as technically and economically feasible, for the purposes of predicting the resting state. That is, our problem possesses significant aleatoric uncertainty (mostly due to chaotic bifurcation dynamics and contact phenomena), but not necessarily significant epistemic uncertainty, and there is no reason to artificially inject such epistemic uncertainty by quantizing the state; rather, a more productive approach might be to measure the continuous state as accurately as possible, and then employ machine learning methods that can work with the full continuous state. 

In particular, we propose to first learn probabilistic models $p=h(s,x,y,\theta,u)=Pr[g(s',x',y',\theta')=True|s,x,y,\theta,u]$ that predict the probability $p$ of bringing the part into a desired configuration by firing solenoid $u$ (and possibly, firing duration) when the part is in configuration $(s,x,y,\theta)$. Here, $(s',x',y',\theta')$ is the successor state resulting from applying the impulse shock. For a multi-step decision policy, it might also be advantageous to explicitly learn a model to predict this state, of the form $(s',x',y',\theta')=f(s,x,y,\theta,u)$. Such a model is known as a forward model in the field of learning control, and if we can learn a sufficiently accurate model of this kind,  we can devise a greedy control policy by choosing the solenoid (and maybe duration) $u^*$ that maximizes the probability of success: $u^*=\argmax_u h(s,x,y,\theta,u)$ \cite{Jordan1992ForwardTeacher,Atkeson1997LocallyControl}.

Learning of the predictive models proceeds in a self-supervised fashion. During training, the system conducts a relatively large number of experimental trials by firing the solenoids randomly and recording the sequence of states by utilizing the overhead computer vision system. In this sequence, the successor state of a trial becomes the starting state of the next trial. Each trial is represented as the tuple 
$(s,x,y,\theta,u,s',x',y',\theta')$. This training data is used together with the success criterion $g(s',x',y',\theta')$ to learn the predictive model $h(s,x,y,\theta,u)$ using a suitable supervised machine learning algorithm.

A small fraction of the  poses are ``leaners" (where the part is leaning on the corral and not flat on the bowl bottom); we simply declare such poses invalid and fire a random impulse to attempt achieving of an acceptable pose.  Although these invalid examples are logged, they are not used for the machine learning inputs.  Finally, the datasets were desk-checked by a human looking at the saved final video frames for quality assurance purposes.  We have found the OpenCV results to be at least 99.8\% accurate.

\section{Manipulation of a Six-Sided Die}\label{sec:die}
To understand the capability of the smart bowl, we attempted a control task where the goal was to rotate a standard six-sided die to a desired configuration with the fewest number of impulses from a random starting position. We developed a simple optical character recognizer to identify which of the die's sides was facing up as shown in Fig. \ref{die_with_states}. This impulse-based manipulation of the die was first explored as a command domain question --- what set of solenoid / impulse pair were actually useful in rolling the die. We designed a sub-task to answer this question.

\subsection{Rolling the Die to any Other Face}
In this sub-task, the goal was to roll the die to any face other than the one it was currently on, easily recognizable by the vision system as a change of the number on the topmost face. Accomplishing this task in a minimal number of attempts is equivalent to maximizing the probability of success in one attempt. When starting in state  $(s,x,y,\theta)$ and ending up in state $(s',x',y',\theta')$, the success criterion is $g(s',x',y',\theta')=True$ iff $s'\neq s$. 

\begin{figure}[!h]
    \centering
    \includegraphics[width=0.48\textwidth]{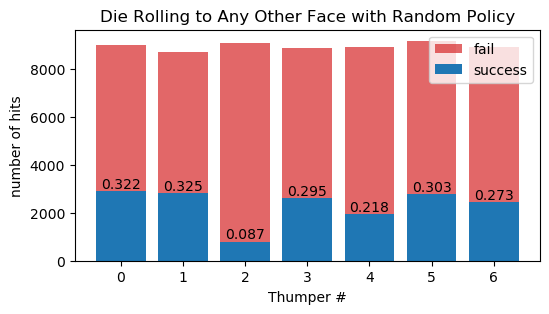}
    \caption{ Die face changing (success) counts and probabilities for each solenoid with a random policy; the average probability of changing the die face with the random policy is 0.260}
    \label{succ_vs_thumper}
\end{figure}

 We started by running an exploratory experiment with 60,000 total firings of the solenoids (requiring about 18 hours of unattended self-supervised operation). Both the solenoid number and the firing duration (impulse of the solenoid) were chosen randomly (firing duration was limited to a 25 millisecond maximum duration). The results are shown as a histogram in Figure \ref{succ_vs_thumper}.  

As the choice of solenoid was random, the total number of hits from each solenoid were roughly the same. However, thumper 2 (the center solenoid) has a much lower chance of succeeding, comparing with other peripheral solenoids. At first, we believed this was a mechanical or electrical defect on that thumper channel, so we physically swapped the center solenoid with a peripheral solenoid, but the low rotation rate remained in the center position. 

Aside from the center solenoid, if we randomly fire any of the peripheral solenoids with a random impulse, there is about 30\% chance to roll the die to another face. This means that there are regions where certain solenoids have little to no authority over the rotation of the die, probably because the solenoid can provide some lift, but not enough rotational "kick" to rotate the die to the next face. 


We then attempted actual control of the die --- finding the optimal solenoid and an effective impulse to provide enough lift to roll the die to any other face. To do so, we need to address the mixed nature of the solenoid impulse mechanism. 

While choice of which solenoid to fire is clearly a categorical choice, the duration of impulse on each solenoid is continuous (at least as viewed on a millisecond scale).  This requires a control policy that can yield simultaneous  multi-class classification (the solenoid number) and a regression-style continuous-valued result (the firing duration). Additionally, our platform is under-actuated, and not all target faces or goal states are reachable from every possible initial location.   Given the speed that sample data is obtainable via the CV system ( $\sim$1 Hz ), we considered data-intensive solutions where seven classifiers (one per solenoid) will be trained.  

Several informal tests were done with a k-nearest neighbors (kNN) classifier \cite{Cover1967NearestClassification}  with k varying from 1 to 24, but the results were not encouraging ---the associated areas under the receiver-operator characteristic (AUROC) curves were on the order of 0.7 at best. On closer inspection, it was found that the sample dataset was strongly biased toward having the die near the edge of the corral, probably due to the die hitting the corral wall and losing energy in the partially inelastic collision. This effect (akin to thermally induced density gradients in a gas) caused significant depletion of the sample population in the bowl center.  In these low density regions, the k-nearest neighborhood diameter was expanding to 10-20mm.  As we found in further testing using a kinematic jig to reproducibly place the die in a controlled location, the regions of the bowl where movements were correlated and consistent are often smaller than 10mm. If those regions happened to be low density as well, then the effective area of the kNN would become much larger than the correlation area and the kNN policy could behave no better than random chance.  

We found significantly better results with a radius neighborhood (rN) classifier; the rN classifier includes all points within a given radius $r$ in the voting set rather than just the $k$ nearest points as in a kNN; voting and final selection of which solenoid to fire proceeds similarly to the kNN and yields the categorical output choosing which solenoid to actuate. The duration of the actuation is then chosen to be the mean of the set of successful activation impulses on that solenoid.

Like kNN, rN relies on some distance metric to determine whether a sample ($x,y,\theta$) is close enough to a prior observation ($x_0, y_0, \theta_0$). As $\theta$ inherits a different unit than $x$ and $y$, we make use of a single distance metric that scalarizes the two distances in position and angle as follows:
\[
D[(x_0,y_0,\theta_0), (x,y,\theta)] = \|(x_0-x, y_0-y)\|_2+w|\theta_0-\theta|
\]
where $w$ is a tuned conversion factor.

\begin{figure}
    \centering
    \includegraphics[width=\linewidth]{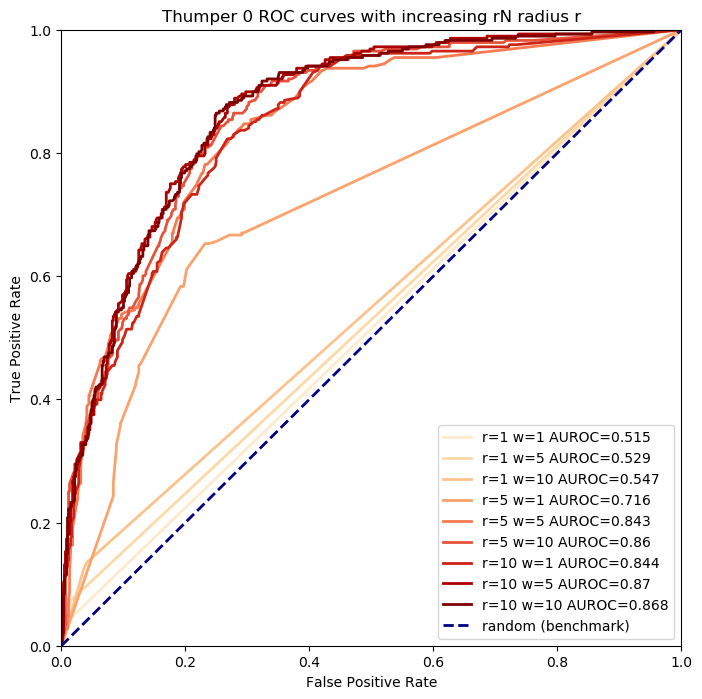}
    \caption{ROC curves for determining $r$ (radius of neighborhood) of the rN classifier; increasing the radius of the neighborhood improves performance but radius beyond 5 mm yields little if any improvement (using an angle conversion factor of $d = \omega \theta  , \omega = 5 mm/deg $) }
    \label{Die_roc}
\end{figure}

To determine an effective radius $r$ and a suitable conversion factor $w$ for the rN classifier, we used 10-fold cross-validation on each of the seven rN classifiers. For each solenoid and its underlying classifier, we took test data from the train-test split and swept a threshold $a$ from 1.0 to 0.0. E.g., when $a=0.7$, for any query in the test split, 70\% of its neighbors (from train split) within $r$ have to meet the success criteria of face different, $g(s',x',y',\theta')=True$ iff $s'\neq s$, for that query to be predicted as successful. Predictions of all queries were then compared with the ground truth label, and an entry of true / false positive rate (TPR / FPR) was plotted. The resulting ROC curves of one of the classifiers is shown in Figure \ref{Die_roc}. The value $r=5$ with $w=5$ was commonly agreed by all seven classifiers from the corresponding AUROC curve value.

\begin{figure}[h!]
    \centering
    \includegraphics[width=\linewidth]{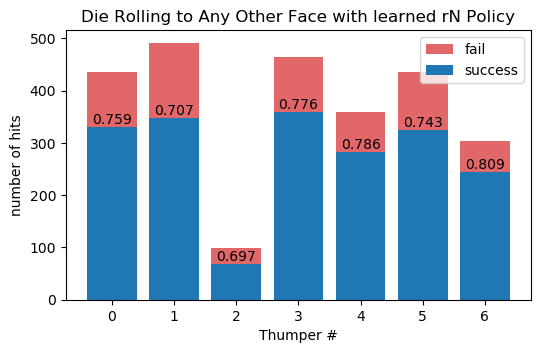}
    \caption{Die face changing (success) counts and rates  on each solenoid using the rN classifier; average is 75.3\%.}
    \label{succ_vs_thumper_with_policy}
\end{figure}

We then performed an rN multi-class classification using these parameter settings and 60,000 verified samples (about 8500 samples per classifier). Given an arbitrary state of the die ($s,x,y,\theta$), the controller will apply all 7 classifiers on this state and then fire the solenoid whose corresponding classifier provides the best probability to roll the die to a different face. The result of this task is shown in Figure \ref{succ_vs_thumper_with_policy} with an average single-shot success probability of 0.753, versus 0.260 for the random policy --- an improvement of almost three times.

\subsection{Controlling the Die to a Specific Target Face}

The next more complicated control problem was to learn how to roll the die to a chosen face different from the one it was currently on, so direction of rolling became significant.

The task here was to achieve a series of 2,000 randomly chosen target values for the upper die face (with no repeated faces) allowing up to 10 impulses to achieve the desired die pose.  This emulates the challenge of feeding properly oriented parts to a manufacturing robot.

\begin{figure}[h!]
    \includegraphics[width=\linewidth]{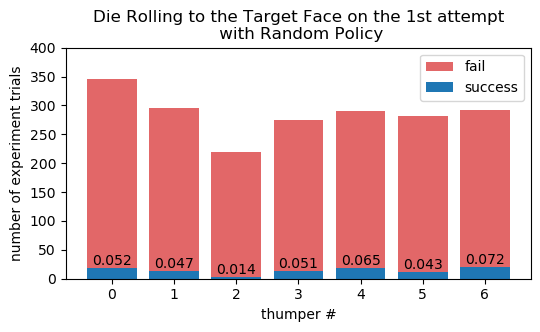}
    \caption{Number of failures (red) and successes (blue) in achieving a targeted goal face for each solenoid as controlled by a random policy. The average success probability is 5.1\% averaged over all thumper channels.} 
    \label{hist_rand}
\end{figure}

The first policy tested was the purely random-choice policy, which served as the experiment's control group. This resulted in an overall 5.1\% success rate for rotating the die to a chosen face.  The per-thumper activations and success rates are shown in Figure \ref{hist_rand}.

\begin{figure}[h!]
    \centering
    \includegraphics[width=\linewidth]{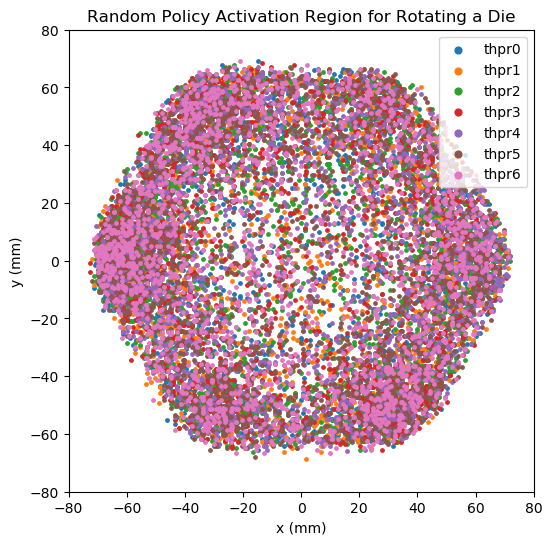}
    \caption{ Random Policy: Die positions and the solenoid fired. } 
    \label{scatter_rand}
\end{figure}

As before, the die's initial position density variation strongly favors the corral wall and avoids the center.  Since this is the random policy (and the experiment control group) we expect to see a uniform distribution of initial positions versus thumper, and we are correct in that (as seen in Figure \ref{scatter_rand}).

We are now in a position to consider a data-driven approach to approximating $h(s,x,y,\theta,u)$ --- the function that predicts the probability $p$  of bringing the part into a desired configuration given the state $(s,x,y,\theta)$ and an impulse $u$; we have the entire 30,000 ground-truth data points for use as the base data for the rN policy.


\begin{figure}[h!]
    \centering
    \includegraphics[width=1\linewidth]{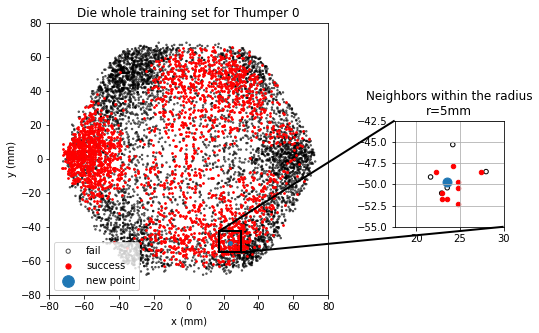}
    \caption{ Training data used in the rN model for classification; inset shows an example of a radius $r$ = 5mm neighborhood of a die at [23, -49, 196] that will be evaluated.}
    \label{Die_neighbors}
\end{figure}

\begin{figure}[!h]
    \centering
    \includegraphics[width=0.9\linewidth]{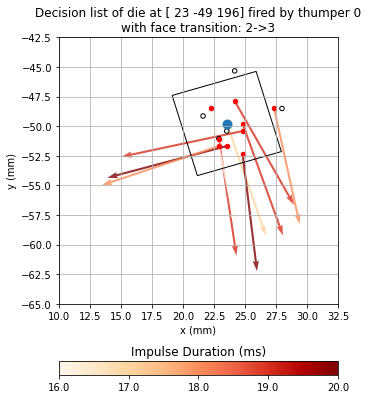}
    \caption{ Sample decision list map of the die from Figure \ref{Die_neighbors} at [23, -49, 196]; the neighbors within the radius $r$ and their rolling directions are shown. }
    \label{die_decision}
\end{figure}

For each of the seven solenoids, we formed a list of all $(s,x,y,\theta)$ examples within the $r=5$ radius.  Based on these training examples, we calculate the success probability for each of the seven solenoids and select the one with the highest success probability.  To determine impulse duration, we  took the mean of the successful impulses for that solenoid. In the case of a tie between two solenoids, we chose one at random from the tied candidates.  An example of the decision neighborhood for a die at [23, -49, 196] is shown in Figure \ref{die_decision}.

\begin{figure}[h!]
    \centering
    \includegraphics[width=\linewidth]{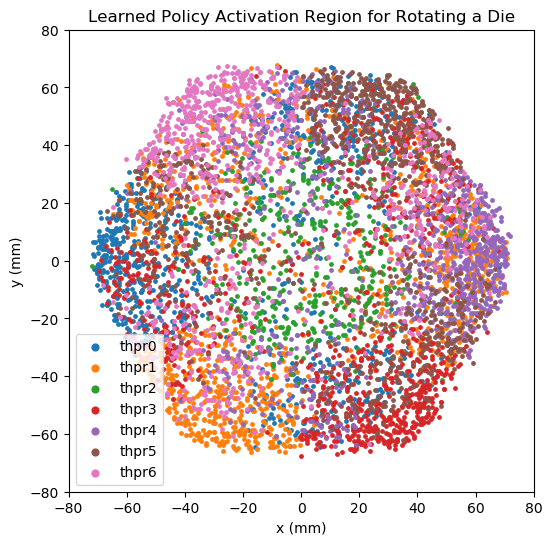}
    \caption{ Die positions and the solenoid fired by the learned rN policy seeking a particular target face. The Voronoi-like segments are impure because the target face varies.} 
    \label{scatter_learned}
\end{figure}

\begin{figure}[h!]
    
    \includegraphics[width=\linewidth]{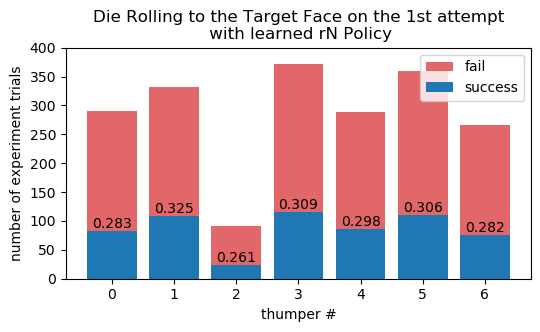}
    \caption{Number of impulses fired and successes. sorted by solenoid; the overall average success rate on the first impulse is 30.6\%. Note the low density on the center solenoid (\#2) is correctly accommodated by the rN policy.} 
    \label{hist_learned}
\end{figure}

As we are only choosing the best option without looking more than one step ahead, we characterize this as a greedy 1-step horizon approach to solve the under-actuated control problem. Using this policy, we ran the same 2,000-random-goals experiment.  Figure \ref{scatter_learned} shows a scattergram of the XY positions of the die, with the color of each dot indicating the particular thumper chosen by the rN policy to have the best chance to rotate the die to another face.  

The final results for this policy are shown in Figure \ref{hist_learned}, with the rN policy achieving the chosen goal state 30.6\%  of the time on the first impulse, beating the benchmark random policy by a factor of $\sim$6 times for single impulses,  and succeeding 97.5\% in 10 or fewer tries, versus the benchmark random policy of 43.0\% in 10 or fewer tries.

\subsection{Controlling the Die with a Two-Step-Horizon
Model Predictive Controller (MPC)}

We then considered examining longer-horizon strategies where the die could be subjected to a series of impulses, with re-evaluation after each impulse, and repeating this strategy until the die is in the goal state, or ten trials have occurred.  This policy is equivalent to a discrete-time two-step-horizon MPC controller when the dynamics model itself is learned from examples. This was done by extracting the $r=$5 mm neighborhood to obtain a set of likely  poses after the first impulse, and performing the greedy one-step-horizon algorithm using each of the first-impulse poses as the initial position.  The probabilities of these 2-impulse final states were calculated as in the one-step-horizon greedy policy, and the weighted sums of these probabilities were rolled back against each of the first-impulse probabilities to calculate success probabilities for the first impulse looking up to two impulses ahead.  As in typical MPC methods, the actions to be taken in the next time step repeat this calculation from the start, and actions postulated on the second and further steps are in no sense "locked in" by the control policy.

\begin{figure}[h!]
    \centering
    \includegraphics[width=1\linewidth]{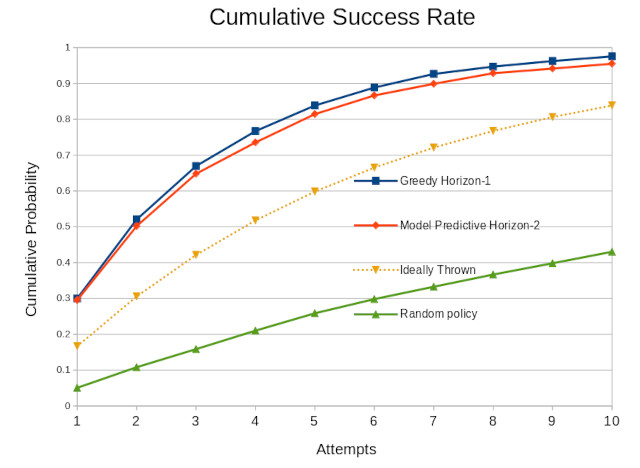}
    \caption{ Cumulative success probabilities to rotate the die to a randomly chosen goal face, with up to 10 attempts allowed.}
    \label{fig:comparison_curves}
\end{figure}

Figure \ref{fig:comparison_curves} summarizes the results of all three policies (random, greedy, and MPC) for the targeted-face task, as well as a purely theoretical ideally thrown die policy.

The horizon-2 MPC controller achieved success in one try at 29.5\% of the time, versus a pure fixed-neighborhood greedy strategy at 30.0\% ; this small difference is  insignificant compared to the random policy of 5.1\% success.  Similarly, the greedy horizon-1 policy achieved a second-shot hit rate of 52.0\% versus 50.2\% for the horizon-2 MPC controller, also an insignificant difference when compared to the random policy at two shot success rate of  $\sim$10.8\% . 

Both the Greedy horizon-1 and MPC horizon-2 were also clearly superior to an idealized random throw of the die (theoretical success rate of 1/6=0.1667).  To be clear, the Thumper mechanism cannot execute random throws the way a human could, and it would be fairly difficult and expensive to generate an ideal die throw with other mechatronic components. For this reason, we do not consider the ideally thrown method as a viable real-world alternative, but even if it were feasible, both the Greedy and MPC controllers are superior.

\section{Discussion and Future Work}\label{sec:conclusion}
The experiments above show that impulse-based manipulation can be effective for object orientation when driven by an ML controller, even when treating the object and impulse manipulator as completely black boxes and with zero modeling of the actual contact physics. That is, it \emph{can} be effective, not necessarily \emph{will} be effective, given the poor showing of the kNN classifier over the rN classifier at 30\%. Thus, the main contribution of this paper is the identification of a relatively less known variant of the kNN classifier --- the rN method --- as a very effective component of a learning controller for part manipulation. The six-fold increase in the success rate of the controller, compared with a random firing policy, and the associated six-fold decrease in the takt time of the system, combined with the minimal need for manual supervision of the method (all training is self-supervised), could possibly result in a very fast and cost-effective method for part manipulation for robotic assembly. 

Future extensions of this work that we are considering include multiple objects in the bowl simultaneously, multiple solenoids being fired simultaneously or with inter-firing delays on the order of the flexure propagation time of the bowl bottom surface, the testing of other shapes, and the integration of the Thumper system with an industrial robot. Although our current best results were from a greedy controller, improved predictive models of system dynamics might lead to superior success rates of MPC schemes with longer horizons, too.


\bibliographystyle{IEEEtran}
\bibliography{DanielsRefsFromMendeley}

\begin{thebibliography}{10}
\providecommand{\url}[1]{#1}
\csname url@samestyle\endcsname
\providecommand{\newblock}{\relax}
\providecommand{\bibinfo}[2]{#2}
\providecommand{\BIBentrySTDinterwordspacing}{\spaceskip=0pt\relax}
\providecommand{\BIBentryALTinterwordstretchfactor}{4}
\providecommand{\BIBentryALTinterwordspacing}{\spaceskip=\fontdimen2\font plus
\BIBentryALTinterwordstretchfactor\fontdimen3\font minus
  \fontdimen4\font\relax}
\providecommand{\BIBforeignlanguage}[2]{{%
\expandafter\ifx\csname l@#1\endcsname\relax
\typeout{** WARNING: IEEEtran.bst: No hyphenation pattern has been}%
\typeout{** loaded for the language `#1'. Using the pattern for}%
\typeout{** the default language instead.}%
\else
\language=\csname l@#1\endcsname
\fi
#2}}
\providecommand{\BIBdecl}{\relax}
\BIBdecl

\bibitem{Sgriccia1950Feeder2654465}
M.~T. Sgriccia, ``{Feeder bowl, US Patent 2654465},'' 1950.

\bibitem{Atkeson1997LocallyControl}
C.~G. Atkeson, A.~W. Moore, and S.~Schaal, ``{Locally weighted learning for
  control},'' \emph{Artificial Intelligence Review}, vol.~11, pp. 75--113,
  1997.

\bibitem{Chung2016PredictableEngines}
S.~Chung and N.~Pollard, ``{Predictable behavior during contact simulation: a
  comparison of selected physics engines},'' \emph{Computer Animation and
  Virtual Worlds}, vol.~27, no. 3-4, pp. 262--270, 2016.

\bibitem{Christiansen1991LearningModels}
A.~D. Christiansen, M.~T. Mason, and T.~M. Mitchell, ``{Learning reliable
  manipulation strategies without initial physical models},'' \emph{Robotics
  and Autonomous Systems}, vol.~8, no. 1-2, pp. 7--18, 1991.

\bibitem{Narendra1974LearningSurvey}
K.~S. Narendra and M.~A.~L. Thathachar, ``{Learning automata - a survey},''
  \emph{IEEE Transactions on systems, man, and cybernetics}, no.~4, pp.
  323--334, 1974.

\bibitem{Ljung1997SystemUser}
L.~Ljung, \emph{{System identification: Theory for the user}}.\hskip 1em plus
  0.5em minus 0.4em\relax Pearson, 1997.

\bibitem{Narendra1990IdentificationNetworks}
K.~S. Narendra and K.~Parthasarathy, ``{Identification and control of dynamical
  systems using neural networks},'' \emph{IEEE Transactions on neural
  networks}, vol.~1, no.~1, pp. 4--27, 1990.

\bibitem{Jordan1992ForwardTeacher}
M.~I. Jordan and D.~E. Rumelhart, ``{Forward models: Supervised learning with a
  distal teacher},'' \emph{Cognitive science}, vol.~16, no.~3, pp. 307--354,
  1992.

\bibitem{Narendra1992NeuralSystems}
K.~S. Narendra and K.~Parthasarathy, ``{Neural networks and dynamical
  systems},'' \emph{International Journal of Approximate Reasoning}, vol.~6,
  no.~2, pp. 109--131, 1992.

\bibitem{Moerland2020Model-basedSurvey}
T.~M. Moerland, J.~Broekens, and C.~M. Jonker, ``{Model-based reinforcement
  learning: A survey},'' \emph{arXiv preprint arXiv:2006.16712}, 2020.

\bibitem{Pfrommer2020Contactnets:Representations}
S.~Pfrommer, M.~Halm, and M.~Posa, ``{Contactnets: Learning discontinuous
  contact dynamics with smooth, implicit representations},'' \emph{arXiv
  preprint arXiv:2009.11193}, 2020.

\bibitem{Parmar2021FundamentalDynamics}
M.~Parmar, M.~Halm, and M.~Posa, ``{Fundamental challenges in deep learning for
  stiff contact dynamics},'' in \emph{2021 IEEE/RSJ International Conference on
  Intelligent Robots and Systems (IROS)}.\hskip 1em plus 0.5em minus
  0.4em\relax IEEE, 2021, pp. 5181--5188.

\bibitem{Cover1967NearestClassification}
T.~Cover and P.~Hart, ``{Nearest neighbor pattern classification},'' \emph{IEEE
  transactions on information theory}, vol.~13, no.~1, pp. 21--27, 1967.

\end{thebibliography}


\begin{thebibliography}{unsrt}

\bibitem{vibratory_initial_patent}
US Patent 2654465 "Feeder Bowl", Mario T Sgriccia, filed 1950-12-09

\bibitem{FreeCAD}
Open source software available at https://www.freecadweb.org/

\bibitem{KiCAD}
Open source software available at https://www.kicad.org/

\bibitem{Zhang-method} 
Z. Zhang "A Flexible New Technique For Camera Calibration", IEEE Transactions on Pattern Analysis and Machine Intelligence | December 2000, Vol 22: pp. 1330-1334, also available at https://www.microsoft.com/en-us/research/project/a-flexible-new-technique-for-camera-calibration-2/publications/

\bibitem{Cover-and-Hart}
Cover, Thomas M.; Hart, Peter E. (1967). "Nearest neighbor pattern classification" IEEE Transactions on Information Theory. 13 (1): 21–27. doi:10.1109/TIT.1967.1053964.

\end{thebibliography}

\end {document}